\pdfoutput=1

\documentclass[11pt]{article}
\usepackage{EMNLP2024}

\usepackage{times}
\usepackage{latexsym}

\usepackage{amsmath}
\usepackage{amsfonts}
\usepackage{graphicx} 
\usepackage{subfigure}
\usepackage{booktabs}
\usepackage{array}
\usepackage{enumitem}
\usepackage{amsthm}
\usepackage{algpseudocode}
\usepackage[switch]{lineno}
\usepackage[utf8]{inputenc}
\usepackage{eqparbox}
\usepackage[nopar]{lipsum}
\usepackage{multirow}
\usepackage{makecell}
\usepackage{xcolor}
\usepackage{hhline} 
\usepackage{microtype}
\usepackage{diagbox}
\usepackage{adjustbox}
\usepackage{tabularx}
\usepackage{multirow}
\usepackage{hyperref}

\usepackage{relsize}

\usepackage{booktabs, multirow, array}
\newcolumntype{N}{@{}m{0pt}@{}}

\usepackage[many]{tcolorbox}
\newtcolorbox{fancyquotes}{%
    enhanced jigsaw, 
    breakable,      
    frame hidden,   
    left=0.5cm,       
    right=0.1cm,      
    overlay={%
        \node [scale=8,
            text=black,
            inner sep=0pt,] at ([xshift=-1cm,yshift=-1cm]frame.north west){}; 
        \node [scale=8,
            text=black,
            inner sep=0pt,] at ([xshift=1cm]frame.south east){};  
            },
                parbox=false,
}

\usepackage{mathtools, nccmath}
\usepackage{scrextend}
\deffootnote[.25in]{.25in}{.15in}{\makebox[.25in][r]{\thefootnotemark .\hspace{.15in}}}

\makeatletter

\newtheorem*{proof*}{Proof}

\usepackage{algorithm}
\usepackage{listings}
\usepackage{color}

\definecolor{codeblue}{rgb}{0.25,0.5,0.5}

\def\@fnsymbol#1{\ensuremath{\ifcase#1\or \dagger\or *\or \ddagger\or
   \mathsection\or \mathparagraph\or \|\or **\or \dagger\dagger
   \or \ddagger\ddagger \else\@ctrerr\fi}}

\newcolumntype{C}[1]{>{\centering\let\newline\\\arraybackslash\hspace{0pt}}m{#1}}

\ExplSyntaxOn
\NewExpandableDocumentCommand { \ValuePlusOne } { m } 
  { \int_eval:n { \int_use:c { c @ #1 } + 1 } }
\NewExpandableDocumentCommand { \Sec } { m } 
  { \fp_eval:n { secd ( #1 ) } }
\NewDocumentCommand { \Rot } { m }
  { 
    \hbox_to_wd:nn { 1 em }
      { 
        \hbox_overlap_right:n 
          { 
            \skip_horizontal:n { \fp_to_dim:n { 7 * cosd (\Angle) } } 
            \rotatebox{\Angle}{#1}
          } 
      } 
  }
\ExplSyntaxOff

\def\Angle{45}
    
\bigskip
\def\Angle{90}

\title{Large Language Models as Foundations for Next-Gen Dense Retrieval: A Comprehensive Empirical Assessment}

\author{
Kun Luo$^{1,2}$\footnotemark[1]\thanks{Equal contribution} \ \
Minghao Qin$^{2}$\footnotemark[1] \ \
Zheng Liu$^{2}$\footnotemark[2]\thanks{Corresponding author} \ \ \
Shitao Xiao$^{2}$ \ \
Jun Zhao$^{1}$ \ \
Kang Liu$^{1,2}$\footnotemark[2]
\\
\textsuperscript{1}{Institute of Automation, Chinese Academy of Sciences} \\
\textsuperscript{2}{Beijing Academy of Artificial Intelligence} \\
{\tt \{luokun695, zhengliu1026\}@gmail.com} \ \ \ 
{\tt kliu@nlpr.ia.ac.cn}
}

\begin{document}
\maketitle
\begin{abstract}
Pre-trained language models like BERT and T5 serve as crucial backbone encoders for dense retrieval. However, these models often exhibit limited generalization capabilities and face challenges in improving in-domain accuracy. Recent research has explored using large language models (LLMs) as retrievers, achieving state-of-the-art performance across various tasks. Despite these advancements, the specific benefits of LLMs over traditional retrievers and the impact of different LLM configurations—such as parameter sizes, pre-training duration, and alignment processes—on retrieval tasks remain unclear.

In this work, we conduct a comprehensive empirical study on a wide range of retrieval tasks, including in-domain accuracy, data efficiency, zero-shot generalization, lengthy retrieval, instruction-based retrieval, and multi-task learning. We evaluate over 15 different backbone LLMs and non-LLMs. Our findings reveal that larger models and extensive pre-training consistently enhance in-domain accuracy and data efficiency. Additionally, larger models demonstrate significant potential in zero-shot generalization, lengthy retrieval, instruction-based retrieval, and multi-task learning. These results underscore the advantages of LLMs as versatile and effective backbone encoders in dense retrieval, providing valuable insights for future research and development in this field.
\end{abstract}

\section{Introduction}
Dense retrieval, a novel paradigm in Information Retrieval (IR), has emerged with the advancement of deep neural networks. Unlike traditional IR methods, dense retrieval encodes both queries and documents as embeddings within a shared latent space, capturing their semantic relationships through embedding similarities. Dense retrieval models have become the predominant choice in recent neural retrieval approaches and are widely applied in various downstream tasks such as web search, question answering, and sentence similarity \cite{karpukhin2020dense,xiong2020approximate,muennighoff2022mteb}.

In the past few years, dense retrieval models intensively adopted pre-trained language models, such as BERT \cite{devlin2018BERT} and T5 \cite{raffel2020exploring}, as their backbone encoders. These models excel in identifying semantic similarities between queries and documents. However, they still face significant challenges in becoming versatile enough to handle a wide range of retrieval tasks \cite{muennighoff2022mteb}. Their in-domain retrieval accuracy is often constrained by the capacity of their backbone encoders, such as the number of parameters \cite{ni2021large}. Additionally, dense retrieval models typically struggle to generalize to unseen data, necessitating fine-tuning with a large amount of labeled data to perform well in the target domain. Finally, achieving versatility in dense retrieval models requires training on multiple retrieval tasks simultaneously, which demands sufficient capacity from the backbone encoder \cite{zhang2023retrieve, bge_embedding}.

Recently Large Language Models (LLMs) have been prompted or fine-tuned as dense retrieval models and achieved improved performance across a wide range of retrieval tasks, thanks to their superior capability for semantic understanding and rich world knowledge \cite{li2023making, wang2023improving, zhuang2024promptreps, muennighoff2024generative}. These models vary in parameters from 2 billion to 56 billion, with pre-training sufficiency ranging from hundreds of billions to tens of trillions of tokens, and include both base models and human preference aligned chat models.
Despite the common understanding that larger models generally yield better performance \cite{kaplan2020scaling, biderman2023pythia}, the specific benefits of varying parameter numbers, pre-training sufficiency, and alignment processes of backbone LLMs for different retrieval tasks still remain unclear.


In this study, we focus on the following two research questions: 1) For different retrieval tasks, what specific benefits can LLMs offer compared to non-LLMs as the backbone encoders? 
2) For LLMs with varying configurations (i.e., different parameter numbers, pre-training sufficiency and alignment processes), what contributes more to different retrieval tasks as the backbone encoder.
We conduct comprehensive empirical investigation across a wide range of retrieval tasks, assessing various critical retrieval capabilities: in-domain accuracy, data efficiency, zero-shot generalization, lengthy retrieval generalization, instruction-based retrieval, and multi-task learning. Our study explore over 15 different backbone LLMs and non-LLMs, with parameter numbers ranging from 0.1 billion to 32 billion and varying pre-training sufficiency, including both base LLMs and chat LLMs.


Previous dense retrieval models have demonstrated inferior \textbf{in-domain accuracy}  due to the limited capacity of their backbone encoders \cite{ni2021large}. We employ MS MARCO \cite{nguyen2016ms}, one of the largest web search datasets, to train and evaluate the in-domain accuracy of dense retrieval models with different backbone encoders. Our results indicate that both increasing the model size and enhancing pre-training sufficiency can consistently improve the upper limit of in-domain accuracy. Notably, we discover that both base LLMs and human-preference-aligned chat LLMs show comparable potential as backbone encoders for dense retrieval tasks. By training with different proportions of MS MARCO, we explore \textbf{data efficiency} and find that scaling up model size facilitates convergence, allowing LLMs to converge swiftly even with limited annotated data, without the need for intricate multi-stage training processes.

We examine generalization ability from three perspectives: zero-shot generalization, lengthy retrieval generalization, and instruction-based retrieval generalization. First, we evaluate \textbf{zero-shot generalization} using BEIR benchmark \cite{thakur2021beir}. Our findings indicate that model size is the most crucial factor for zero-shot retrieval generalization. Moreover, traditional dense retrieval models are limited by the maximum input length used during pre-training and retrieval training. We investigate whether LLM-based retrievers, pre-trained with longer context windows, can effectively generalize to \textbf{lengthy retrieval} tasks even when trained with shorter passage lengths. Finally, dense retrieval models often lack flexibility in handling varying retrieval intents \cite{su2022one}. We explore the capability of different models to \textbf{incorporate instructions} during retrieval, discovering that training with instruction benefits LLMs but not non-LLMs, and that human-preference alignment does not significantly improve performance compared to base LLMs.

We further explore the \textbf{multi-task learning} capabilities of models with different backbone encoders, essential for developing versatile retrievers \cite{zhang2023retrieve, bge_embedding}. We adopt five distinct retrieval tasks, where interference exists due to varying retrieval intents. Our findings reveal that although all models experience performance decreases with multi-task training compared to training on each single-task, increasing model size consistently mitigates this gap.

To summarize, we make the following contributions: 
1) We conduct a thorough experimental study using more than 15 backbone encoders with different configurations for dense retrieval across six distinct retrieval tasks.
2) We demonstrate that LLM-based retrievers consistently enhance performance across all retrieval tasks compared to non-LLM-based retrievers.
3) We investigate how different configurations of backbone LLMs impact each retrieval task, focusing on distinct retrieval capabilities.

\section{Related Work}
The related works are reviewed from two aspects: dense retrieval, LLM-based retriever.

First of all, in the realm of neural retrievers, dense retrieval models have consistently demonstrated superior performance over traditional sparse models like BM25 across a wide array of retrieval tasks \cite{karpukhin2020dense, ni2021large, muennighoff2022mteb}. A critical factor contributing to the success of dense retrieval models is the utilization of powerful pre-trained language models as their initialization.

Over the past few years, pre-trained language models such as BERT \cite{devlin2018BERT} and T5 \cite{raffel2020exploring} have been intensively used as backbone encoders for dense retrieval. For instance, GTR \cite{ni2021large} highlights the in-domain accuracy and generalization capabilities of T5-based dense retrieval models, with model parameters reaching up to 4.8 billion.  \citet{fang2024scaling} explores scaling laws for dense retrieval models but restricts their study to BERT backbones with up to 110 million parameters and only explores the in-domain situation. Currently, state-of-the-art dense retrievers employ models with more than 7 billion parameters or more as backbones.
 \citet{neelakantan2022text} discuss large-scale unsupervised text embedding pre-training, observing consistent performance improvements when scaling up GPT-based dense retrieval model sizes from 300 million to 175 billion parameters. Additionally, recent studies such as  \citet{wang2023improving} have shown that fine-tuning directly with labeled data can achieve strong performance. Our study focuses on fine-tuning directly using labeled data while comparing various backbone encoders.

Large Language Models (LLMs) have recently demonstrated significant potential as backbone encoders for dense retrieval, attributed to their vast number of parameters and extensive pre-training.
Repllama \cite{ma2023fine} fine-tuned Llama-2-7B and Llama-2-13B to function both as dense retrievers and pointwise rerankers. LLaRA \cite{li2023making} introduced two pretraining tasks specifically designed to better adapt the backbone Llama-2-7B model for dense retrieval, resulting in notable improvements in both supervised and zero-shot scenarios. E5-mistral and Gecko \cite{wang2023improving, lee2024gecko} enhanced the training of LLM-based dense retrievers using synthetic data, employing models with 1.5 billion and 7 billion parameters to achieve notable results across various retrieval tasks. GRIT \cite{muennighoff2024generative} successfully unified text embedding and generation within a single LLM, maintaining performance levels comparable to those of specialized embedding-only and generative-only models, using a model with 56 billion parameters (14 billion activation parameters). LLM2Vec \cite{behnamghader2024llm2vec} presented an unsupervised method for transforming decoder-only LLMs into dense retrievers, demonstrating significant promise for adapting LLM backbone encoders for dense retrieval in an unsupervised manner. PromptReps \cite{zhuang2024promptreps} employed human preference-aligned chat LLMs to produce high-quality dense representations unsupervised.

These models vary in parameters from 1.5 billion to 56 billion, with pre-training covering hundreds of billions to tens of trillions of tokens, and include both base LLMs and human preference-aligned chat LLMs. Despite the exciting advancements in retrieval tasks achieved by leveraging various LLMs with distinct configurations and diverse training strategies, the specific benefits of variations in parameter count, pre-training extent, and alignment processes of backbone LLMs for retrieval tasks remain still uncertain.

\section{Preliminary}
Dense retrieval leverages an encoder to project both the query $\mathrm{q}$ and the candidate passage $\mathrm{p}$ into a shared dense embedding space, resulting in embeddings $\mathrm{h}_q$ and $\mathrm{h}_p$. A scoring function, such as the inner product or cosine similarity, is then applied to these dense vectors to model relevance:
\begin{equation}
    \mathrm{s}(\mathrm{q}, \mathrm{p}) = \langle \mathrm{h}_q,\mathrm{h}_p \rangle
\end{equation}
This allows for the retrieval of relevant documents by performing approximate nearest neighbor (ANN) search within the embedding space.

In our study, we compare more than 15 backbone encoders, varying in model architecture (encoder-only and decoder-only), model size (0.1B to 32B), and pre-training sufficiency (up to 15T tokens). Consistent with prior research, we utilize the $\mathrm{[CLS]}$ token to obtain text representations for the BERT model and employ mean-pooling for the T5 model. For instance, BERT tokenizes the input text into a sequence $\mathrm{T}$: $\mathrm{[CLS]}$, $\mathrm{t}_1$, ..., $\mathrm{t}_N$, $\mathrm{[EOS]}$. This tokenized sequence is subsequently encoded by BERT, generating output embeddings that are combined to form the text embedding, with the $\mathrm{[CLS]}$ token performing this integration: 
\begin{align}
    \mathrm{h}_t &= \mathrm{BERT}(\mathrm{T})[\mathrm{CLS}] \label{eq:cls}
\end{align}
When using large language model (LLM) as the backbone encoder, text embeddings need to be created differently. Most LLMs use a decoder-only architecture and causal attention mechanism, meaning that only the last token in the input sequence can access the global context. As a result, the text embedding is taken from the output embedding of the special token $\mathrm{[EOS]}$:
\begin{equation}
    \mathrm{h}_t = \mathrm{LLM}(\mathrm{T})[\mathrm{EOS}]
\end{equation}
Given the query-passage pair $(\mathrm{q}_i, \mathrm{p}_i^+)$, we adopt the standard InfoNCE \cite{izacard2021unsupervised} loss $\mathrm{L}$ over the in-batch negatives and hard negatives for training:
\begin{equation}
    \mathrm{L} = - \lg \frac{\exp(\mathrm{s}(\mathrm{q}_i, \mathrm{p}_i^+))}{
    \exp(\mathrm{s}(\mathrm{q}_i, \mathrm{p}_i^+)) + \sum\limits_{j} \exp(\mathrm{s}(\mathrm{q}_j, \mathrm{p}_j^-))}
    \label{eq:loss_function}
\end{equation}
where ${\mathrm{p}_j^-}$ is the set of negative passages and $\mathrm{s}(\mathrm{q}, \mathrm{p})$ is the scoring function of query and passage. In this paper, we adopt the temperature-based cosine similarity function as follows:
\begin{equation}
    \mathrm{s}(\mathrm{q}, \mathrm{p}) = \frac{1}{\tau}\cos(\mathrm{h}_q, \mathrm{h}_p)
\end{equation}
$\tau$ is a temperature hyper-parameter, which is fixed to 0.02 in all experiments.

\begin{figure*}[t]
\centering
\includegraphics[width=0.97\linewidth, height=0.39\linewidth]{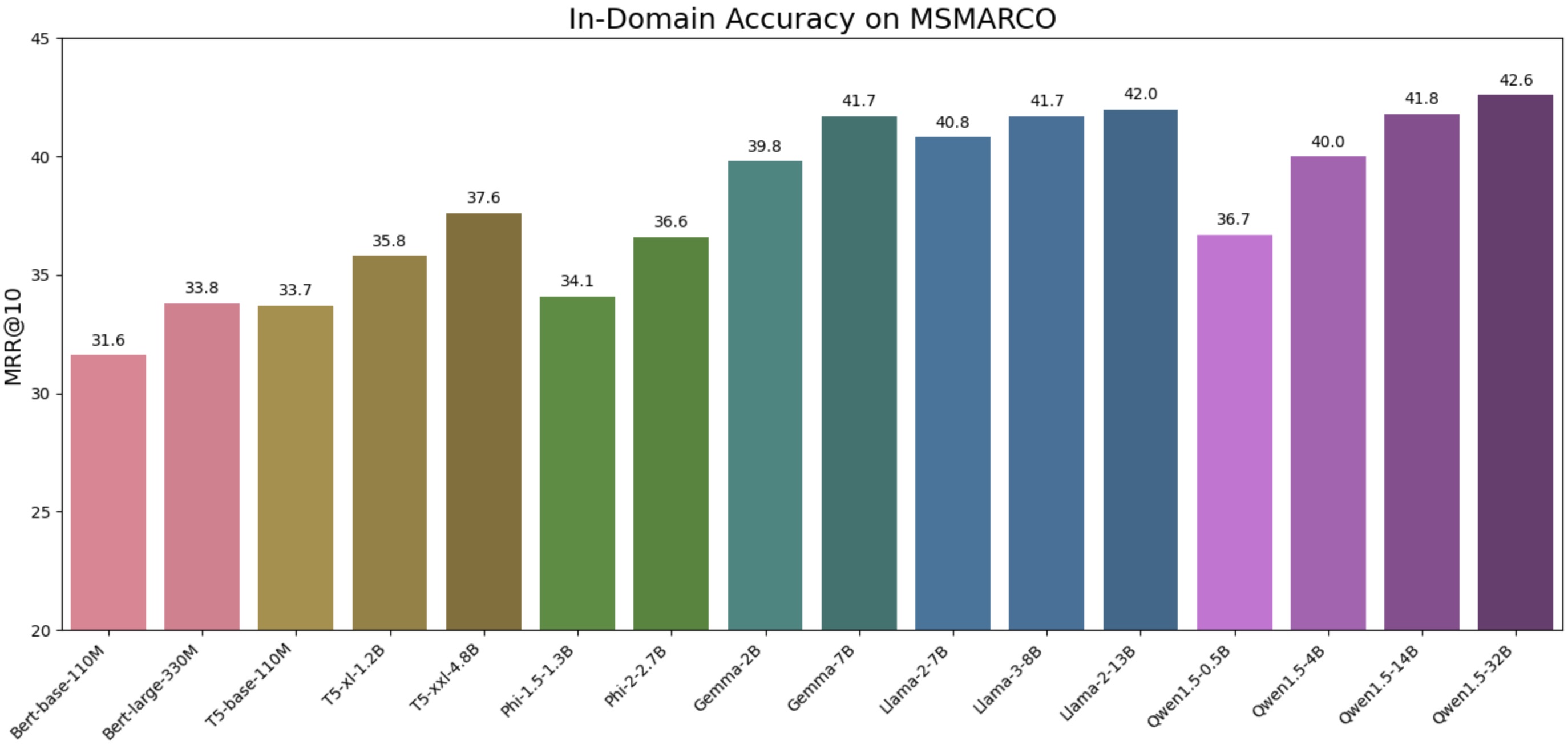}
\vspace{-12pt}
\caption{In-domain accuracy (measured by MRR@10)}
\vspace{-16pt}
\label{fig:accuracy}
\end{figure*}

\vspace{-3pt}

\begin{figure}[t]
\centering
\centering
\includegraphics[width=0.97\linewidth,  height=0.612\linewidth]{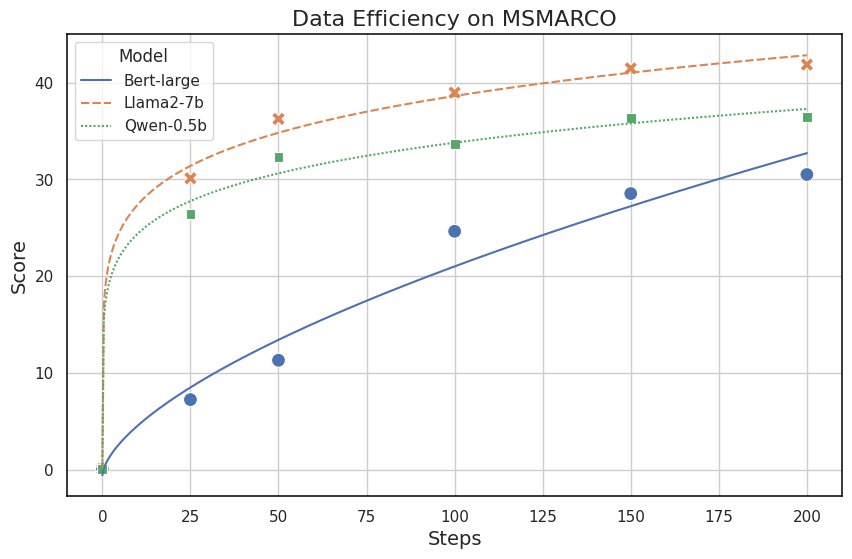}
\vspace{-7pt}
\caption{Data efficiency}
\vspace{-15pt}
\label{fig:DataEfficiency}
\end{figure}

\section{Empirical Study}

In this section, we aim to address two key research questions: 
1) For different retrieval tasks, what specific benefits can LLMs offer compared to non-LLMs as the backbone encoders?
2) For LLMs with varying configurations (i.e., different parameter numbers, pre-training sufficiency, and alignment processes), what contributes more to different retrieval tasks as the backbone encoder. 
To answer these questions, we conduct a comprehensive empirical study across six critical dimensions of dense retrieval, each encompassing several specific retrieval tasks. These dimensions are investigated using various pre-trained language models as backbone encoders, focusing on: in-domain accuracy (Section~\ref{In-domain Accuracy}), data efficiency (Section~\ref{Data Efficiency}), zero-shot generalization (Section~\ref{Zero-Shot Generalization}), lengthy retrieval generalization (Section~\ref{Lengthy Retrieval Generalization}), instruction-based retrieval (Section~\ref{Instruction-Based Retrieval}), and multi-task learning (Section~\ref{Multi-Task Learning}).

\subsection{In-domain Accuracy}
\label{In-domain Accuracy}
\textbf{Setting} We utilize MS MARCO \cite{nguyen2016ms} to train and evaluate the in-domain accuracy of dense retrieval models with varying backbones encoders. Specifically, we employ BERT \cite{devlin2018BERT} with 110M and 330M parameters (BERT-base and BERT-large), T5 \cite{raffel2020exploring} encoders with parameter numbers ranging from 110M to 4.8B, and a diverse set of LLMs including the Llama, Phi, Gemma, and Qwen1.5 series  \cite{touvron2023llama, gunasekar2023textbooks, bai2023qwen, team2024gemma}. It is important to note that different LLMs have varying configurations. For instance, the phi-1.5 model is a lightweight LLM with 1.3B parameters and is pre-trained on a relatively small amount of tokens (150B), indicating less pre-training sufficiency. In contrast, the Llama-3-8B model is extensively pre-trained on over 15T tokens, significantly more than the 2T tokens used for Llama-2-7B. The Qwen1.5 series offers a variety of models in different sizes, all pre-trained on the same corpus, enabling direct comparisons of the effects of scaling up model size.

All models are trained with a batch size of 128 and incorporate 7 hard negative samples to ensure fair comparisons of in-domain retrieval accuracy. All training operations take place on 8xA800 (80GB) GPUs. We use the Adam optimizer with an initial learning rate of 3e-4 and linear decay. For training LLM retrievers, we employ LoRA \cite{hu2021lora}, which has demonstrated similar efficacy to full-parameter fine-tuning for retrieval tasks  \cite{ma2023fine}.
The in-domain accuracy of each model is evaluated using the MS MARCO development set, comprising 6,980 queries. We use NDCG@10, MRR@10, Recall@10, and Recall@1000 as evaluation metrics, providing a comprehensive analysis of in-domain performance.

\noindent \textbf{Results and Analysis} As presented in Figure \ref{fig:accuracy}, the results indicate that model performance generally improves with an increase in parameter numbers. This trend is particularly noticeable within models from the same series. For instance, the Qwen1.5 series demonstrates this progression: Qwen1.5-0.5B model scores 36.7, while the Qwen1.5-32B model achieves 42.6, representing an improvement of 5.9 points. This trend suggests that increasing model size is a feasible way to yield better in-domain accuracy. Detailed results are presented in Table \ref{tab:accuracy}.

Additionally, the results demonstrate that LLM-based retrievers significantly outperform non-LLM retrievers. The performance of Gemma-2B has already surpassed all BERT and T5-based models despite having fewer parameters than the T5-xxl model. This suggests that LLMs' extensive pre-training and advanced language understanding capabilities offer significant advantages as backbone encoders for dense retrieval.

An interesting observation is that smaller models can sometimes marginally outperform larger ones. The Qwen1.5-0.5B model, with fewer parameters, surpasses the Phi-1.5-1.3B model and competes closely with the Phi-2-2.7B model. This performance discrepancy may be attributed to differences in pre-training sufficiency. The Qwen1.5 models benefit from more extensive and diverse pre-training data, totaling over 3 trillion tokens, whereas the Phi models are pre-trained on a smaller amount of high-quality data, with 150 billion tokens for the Phi-1.5 and 1.4 trillion tokens for the Phi-2. This extensive pre-training enables the Qwen1.5-0.5B model to perform better when fine-tuned for retrieval tasks.
A similar conclusion can be drawn from the comparison between the Llama-3-8B and Llama-2-7B models, as well as between LLMs and non-LLMs. Extensive and varied pre-training of backbone encoders can significantly enhance in-domain retrieval accuracy, even compensating for a smaller parameter count.

\subsection{Data Efficiency}
\label{Data Efficiency}

\textbf{Setting} We use checkpoints from models trained on MS MARCO for different numbers of steps to evaluate their performance on the development set, in order to better understand the impact of parameter number and pre-training sufficiency on data efficiency and convergence speed.

We compare BERT-large, Qwen1.5-0.5B, and Llama-2-7B to explore the impact of data efficiency with model parameter number and pre-training sufficiency. Notably, BERT-large and Qwen1.5-0.5B have similar non-embedding parameter number, while Qwen1.5-0.5B is based on decoder architecture and has undergone more extensive pre-training.

\begin{figure}[t]
\centering
\includegraphics[width=0.95\linewidth,  height=0.60\linewidth]{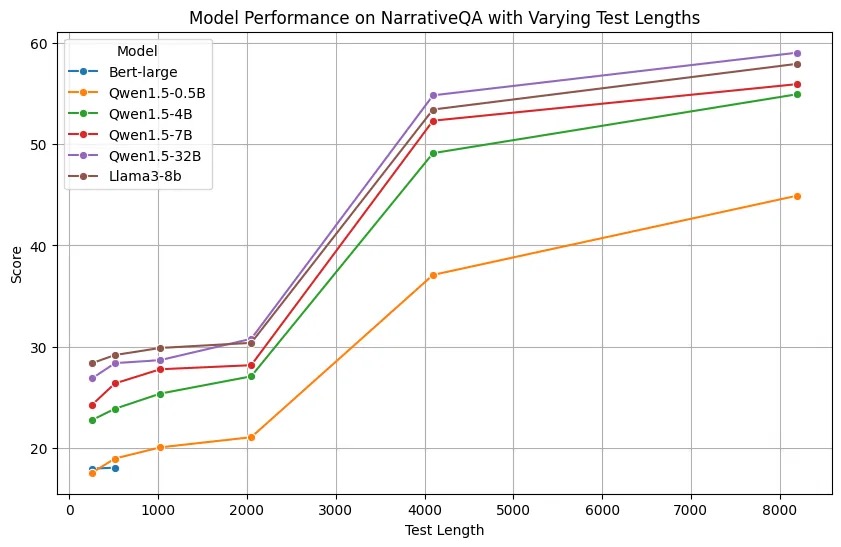}
\vspace{-8pt}
\caption{Lengthy retrieval}
\vspace{-18pt}
\label{fig:LengthyInput}
\end{figure}

\begin{figure*}[t]
\centering
\includegraphics[width=\linewidth, height=0.43\linewidth]{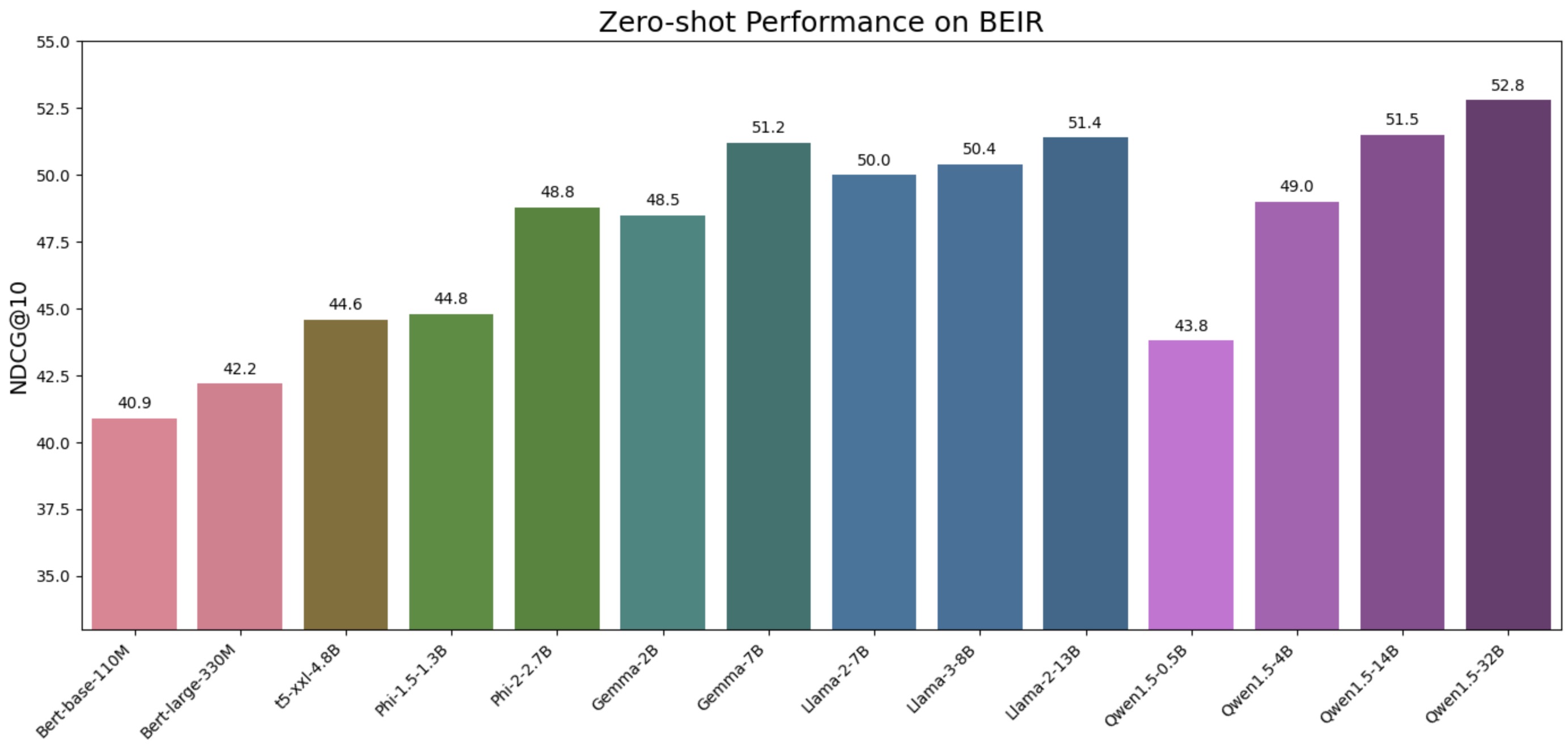}
\vspace{-26pt}
\caption{Zero-shot performance (measured by NDCG@10)}
\vspace{-10pt}
\label{fig:generalization}
\end{figure*}

\noindent \textbf{Results and Analysis} As presented in Figure \ref{fig:DataEfficiency}, our findings indicate that larger model sizes lead to higher data efficiency and faster convergence. Specifically, after 100 training steps on MS MARCO, Llama-2-7B outperforms Qwen1.5-0.5B by 5.4 points and BERT-large by 14.4 points. This suggests that with an increase in parameter number, better performance can be achieved with less labeled data. 
Furthermore, as shown in Table \ref{tab:convergence speed}, when comparing the relative score difference between 100 steps and the full training of 3700 steps, Llama-2-7B shows a score difference of 8.8 points, which is smaller than the 9.7 points for Qwen1.5-0.5B and 15.3 points for BERT-large. This indicates that larger models are able to converge faster.

The experiment results also demonstrate that LLMs have better data efficiency compared to non-LLMs, even with similar parameter sizes. For example, after 100 training steps on MS MARCO, Qwen1.5-0.5B outperforms BERT-large by 9 points. Despite having a similar number of parameters, Qwen1.5-0.5B has undergone more extensive pre-training (over 3 trillion tokens compared to BERT's 3.3 billion tokens) and employs a decoder architecture, which enhances its language understanding ability and enables faster convergence in the retrieval task where text discriminative ability is crucial.

\subsection{Zero-Shot Generalization}
\label{Zero-Shot Generalization}
\textbf{Setting} Dense retrieval models typically struggle with zero-shot retrieval on unseen data \cite{ni2021large}. We investigate the specific benefits that LLM-based retrievers can bring to zero-shot generalization, focusing on varying model sizes and pre-training sufficiency.

We evaluate all models on 13 zero-shot retrieval tasks in the BEIR \cite{thakur2021beir} evaluation suite, which encompasses a diverse range of retrieval tasks and domains, including medical retrieval, financial retrieval, and duplication detection. All models are directly transferred for zero-shot evaluation on BEIR after being trained on MS MARCO. During the evaluations, we set the maximum length of the query to 64 tokens and the maximum length of the passage to 256 tokens.

\begin{table}[t]
    \centering
    \begin{adjustbox}{width=0.48\textwidth}
    \begin{tabular}{lcccc}
    \toprule
    Model & Parameter Number & NDCG@10 & MRR@10 & Recall@10 \\
    \midrule
    \multicolumn{5}{c}{100 Steps} \\
    \midrule
    Bert-large & 0.3 B & 24.6$(\delta=15.3)$ & 20.0 & 40.5 \\
    Qwen1.5-0.5B & 0.5 B & 33.6$(\delta=9.7)$ & 27.9 & 53.2 \\
    Llama-2-7B & 7  B & 39.0$(\delta=8.8)$ & 32.4 & 61.0 \\
    \midrule
    \multicolumn{5}{c}{Full 3700 Steps} \\
    \midrule
    Bert-large & 0.3 B & 39.9 & 33.8 & 60.3 \\
    Qwen1.5-0.5B & 0.5 B & 43.3 & 36.7 & 65.5 \\
    Llama-2-7B & 7  B & 47.8 & 40.8 & 70.9 \\
    \bottomrule
    \end{tabular}
    \end{adjustbox}
    \vspace{-6pt}
    \caption{Model convergence speed.}
    \vspace{-18pt}
    \label{tab:convergence speed}
\end{table}

\noindent \textbf{Results and Analysis} The results are shown in Figure \ref{fig:generalization}, measured by average performance of NDCG@10 across 13 retrieval tasks. LLM retrievers significantly outperform non-LLM retrievers in zero-shot retrieval tasks, indicating that the extensive knowledge and robust generalization capabilities of LLMs are highly advantageous for zero-shot retrieval. Notably, this improvement is not merely a result of increased model size: even the Qwen1.5-0.5B model, which has a similar non-embedding parameter count, demonstrates much better generalization (+1.6\%) than the BERT-large model. This highlights the potential of LLMs to serve as robust encoders for various retrieval domains.

For different configurations of LLMs, model size is the primary factor influencing their generalization capability. Unlike in-domain accuracy, where both model size and pre-training sufficiency are important, generalization performance is almost directly correlated with the number of parameters. For example, the Qwen-0.5B model, despite benefiting from more extensive pre-training, performs worse than the Phi-1.5-1.3B and Phi-2-2.7B models with larger parameter sizes but less pre-training sufficiency. This suggests that larger models, with better capacity, can prevent overfitting to domain-specific retrieval data, resulting in better generalization to unseen data.

\begin{table}[t]
    \centering
    \begin{adjustbox}{width=0.48\textwidth, height=0.068\textwidth}
    \begin{tabular}{lccc}
    \toprule
    Model & Parameter Number & MSMARCO-ID & MSMARCO-OOD \\
    \midrule
    Bert-large & 0.3 B & 40.0 & 39.3 \\
    Qwen1.5-0.5B & 0.5 B & 43.5 & 43.6 \\
    Qwen1.5-4B & 4 B & 47.0 & 47.0 \\
    Qwen1.5-14B & 14 B & 48.9 & 48.9 \\
    Llama-3-8B & 8 B & 49.6 & 49.6 \\
    \bottomrule
    \end{tabular}
    \end{adjustbox}
    \vspace{-5pt}
    \caption{Unseen instruction comparison. "ID" means instructions are seen during training, "OOD" means the instructions are unseen during training.}
    \vspace{-11pt}
    \label{tab:ID and OOD instruction}
\end{table}

\subsection{Lengthy Retrieval Generalization}
\label{Lengthy Retrieval Generalization}

\begin{table*}[t]
    \centering
    \setlength{\tabcolsep}{15pt}
    \begin{adjustbox}{width=0.98\textwidth}
    \begin{tabular}{lccccccc}
    \toprule
     \textbf{Model} &  \textbf{Hotpot} &  \textbf{NQ} &  \textbf{MSM} &  \textbf{FiQA} &   \textbf{NFCorpus} &  \textbf{SciFact} &  \textbf{Average} \\
    \midrule
    BERT-large     &  46.8(-4.6) &  47.3(+0.9) &  40.0(+0.1) &  24.3(-2.0) &  24.7(-2.0) &  55.5(+0.9) &  39.8(-1.0) \\
     Qwen1.5-0.5B   &  59.3(+2.7) &  50.5(+7.1) &  43.5(+0.2) &  33.5(-0.4) &  31.8(+1.5) &  66.2(-0.6) &  47.4(+1.7) \\
     Qwen1.5-4B     &  63.6(-0.1) &  57.7(+7.4) &  47.0(+0.2) &  39.8(+0.4) &  34.8(-0.6) &  72.1(+1.3) &  52.5(+1.4) \\
     Qwen1.5-14B    &  69.5(+3.2) &  63.0(+3.7) &  48.9(+0.2) &  45.6(+0.6) &  37.0(+0.6) &  75.9(+1.7) &  56.7(+1.8) \\
     Llama-3-8B     &  70.9(+4.9) &  63.1(+6.7) &  49.6(+0.9) &  44.8(+3.1) &  37.8(+2.6) &  75.4(+1.4) &  56.8(+3.2) \\
     Qwen1.5-0.5B-Chat   &  57.5 &  49.5 &  43.6 &  32.8 &  31.7 &  65.0 &  46.7 \\
     Qwen1.5-4B-Chat   &  64.0 &  58.1 &  47.2 &  40.2 &  36.1 &  71.3 & 52.8 \\
     Qwen1.5-14B-Chat   &  69.4 &  63.5 &  49.0 &  44.4 &  37.1 &  76.0 &  56.6 \\
     Llama-3-8B-Chat   &  70.6 &  63.0 &  49.6 &  44.8 &  38.2 &  75.5 &  56.9 \\
    \bottomrule
    \end{tabular}
    \end{adjustbox}
    \vspace{-6pt}
    \caption{Instruction-based retrieval performance measured by NDCG@10. The average performance discrepancy is compared to training without instruction.}
    \vspace{-2pt}
    \label{tab:Instruction-based retrieval}
\end{table*}

\begin{table*}[t]
    \centering
    \setlength{\tabcolsep}{22pt}
    \begin{adjustbox}{width=0.98\textwidth}
    \begin{tabular}{lcccccc}
    \toprule
    \textbf{Model} & \textbf{Hotpot} & \textbf{STS}   & \textbf{MSM} & \textbf{Tool}  & \textbf{QReCC} & \textbf{Average} \\
    \midrule
    BERT-large & 62.1(-2.4) & 80.2(+2.7) & 38.8(-1.1) & 76.6(-5.2) & 47.3(-4.1) & 61.0(-2.0) \\
    Qwen1.5-0.5B & 72.1(-1.5) & 80.1(+1.0) & 43.7(+0.2) & 84.8(-4.8) & 50.7(-3.9) & 66.3(-1.8) \\
    Qwen1.5-4B & 79.8(-0.6) & 82.0(+2.2) & 46.8(+0.0) & 
    86.1(-4.2) & 54.9(-4.4) & 69.9(-1.4) \\
    Llama-3-8B & 85.7(+0.3) & 82.8(+1.3) & 48.9(+0.2) &
    89.9(-2.7) & 59.5(-3.3) & 73.4(-0.8) \\
    \bottomrule
    \end{tabular}
    \end{adjustbox}
    \vspace{-6pt}
    \caption{Multi-task learning performance measured by NDCG@10. The performance discrepancy is compared to training on each single task.}
    \vspace{-16pt}
    \label{tab:Multi-task learning}
\end{table*}

\textbf{Setting} Traditional dense retrieval models are constrained by the maximum input length used during pre-training and retrieval training, while extending this length significantly increases computational costs \cite{chen2024bge}. Given that LLMs are pre-trained with longer context windows, we investigate if they can be trained with shorter passage lengths while effectively generalizing to longer lengths during retrieval.
We use MS MARCO for training and set the maximum query length to 64 tokens and the maximum passage length to 256 tokens. All other hyperparameters are aligned with those used in Section \ref{In-domain Accuracy}. 

For evaluation, we utilize NarrativeQA \cite{kovcisky2018narrativeqa}, which requires long context information to accurately retrieve target queries. The evaluation was conducted with maximum lengths ranging from 256 to 8192 tokens for passages, with the goal of thoroughly assessing each model's length generalization capabilities in the retrieval task.

\noindent \textbf{Results and Analysis} The results are illustrated in Figure \ref{fig:LengthyInput}. The long context window of LLMs improves length generalization compared to BERT. When evaluated with a context length of 256 tokens on the NarrativeQA Retrieval task, BERT-large outperforms Qwen1.5-0.5B by 0.4 points. However, with a length of 512 tokens, Qwen1.5-0.5B exceeds the performance of BERT-large by 0.9 points. This interesting finding demonstrates that LLM retrievers consistently generalize better with increasing input lengths, while non-LLM retrievers like BERT struggle with longer inputs and are constrained by a 512-token limit unless explicitly extended. Detailed results are presentend in Table \ref{tab:Lengthy retrieval}

Furthermore, increasing the parameter number of LLM retrievers consistently enhances performance with longer inputs. This indicates that scaling up LLMs is an effective strategy for improving lengthy retrieval generalization, obviating the need for specific training on longer retrieval inputs.

\subsection{Instruction-Based Retrieval}
\label{Instruction-Based Retrieval}
\textbf{Setting} Dense retrieval models often lack flexibility in adapting to varying retrieval intents of users, which is both common and critical in real-world retrieval scenarios \cite{su2022one}. We incorporate instructions into the training of dense retrieval models, aiming to evaluate the instruction comprehension capabilities of models with different backbone encoders. Specifically, we prepare five retrieval instructions and prepend them to queries during training on MS MARCO. We conduct evaluation on six retrieval tasks, including both in-domain and out-of-domain scenarios, to determine whether incorporating instructions can enhance the understanding of retrieval intent thus improving general performance of different models. The instructions are presented in Figure \ref{fig:instructions}.

\noindent \textbf{Results and Analysis} As shown in Table \ref{tab:Instruction-based retrieval}, training with instructions significantly improves the performance of LLM retrievers, whereas for BERT retrievers results in decreased performance. This suggests that LLMs have superior semantic understanding, enabling them to adjust retrieval objectives based on instructions.

We evaluate models on MS MARCO \cite{nguyen2016ms} development set using instructions not seen during training. The result is presented in Table \ref{tab:ID and OOD instruction}. These instructions are complex modifications of the training instructions (Figure \ref{fig:instructions}), designed to test the models' robustness. The results show that LLM retrievers exhibit strong robustness to these new instructions, while BERT experience performance degradation due to interference from the unseen instructions. This implies that LLMs can better utilize their capabilities in real-world retrieval scenarios as backbone encoder for dense retrieval, offering better customizability and adaptability to meet diverse user retrieval needs.

Furthermore, we adopt chat LLMs as backbone encoders to investigate if these aligned models could better utilize retrieval instructions, the result is shown in Table \ref{tab:Instruction-based retrieval}. Contrary to expectations, chat LLMs do not show further improvements when trained and tested under the same setting as base models. Thus, given the superior scalability of base LLMs across various downstream tasks, the base LLMs remain more suitable as backbone encoders for dense retrieval models.

\subsection{Multi-Task Learning}
\label{Multi-Task Learning}
\textbf{Setting} Training a versatile dense retrieval model is challenging due to the specific semantic information required by various retrieval tasks, often causing mutual interference  \cite{zhang2023retrieve, bge_embedding,neelakantan2022text}. We explore the multi-task learning capacity of different backbone encoders, which is essential for developing robust retrievers.

Our study encompasses four distinct retrieval tasks alongside a text similarity task: 1) ToolLLM \cite{qin2023toolllm}: This task evaluates the ability of retrievers to identify necessary tools based on provided instructions and tool descriptions. Performance is measured using NDCG@5 on the test set. 2) QReCC \cite{anantha2020open}: This task involves retrieving relevant knowledge based on the concatenation of conversation context and the most recent query. Performance is assessed using NDCG@3, in line with previous studies \cite{mao2023learning}. 3) NLI \cite{bowman2015large}: We utilize the NLI training set to establish text similarity capabilities and evaluate models on STS tasks from the MTEB \cite{muennighoff2022mteb}. 4) HotpotQA \cite{yang2018hotpotqa}: This task tests retrieval performance in a multi-hop question-answering scenario. 5) MS MARCO \cite{nguyen2016ms}: This task assesses the web search capabilities of different models.

\noindent \textbf{Results and Analysis} As shown in Table \ref{tab:Multi-task learning}, the results demonstrate a clear trend: as model size increases, the average performance across the five distinct retrieval tasks improves. This indicates that larger models exhibit enhanced universality and capacity, suggesting their greater potential to serve as versatile embedding models in multi-task scenarios.

In addition to comparing the absolute performance of each model across multiple tasks, we conducted experiments contrasting the performance of models trained on each individual task versus joint multi-task training. Table \ref{tab:Multi-task learning} presents the relative performance discrepancy. We observed that multi-task training results in a relative performance decrease compared to single-task training across all tasks. This aligns with the hypothesis proposed by  \cite{neelakantan2022text}, suggesting that certain retrieval tasks might have inherently conflicting definitions, such as search and sentence similarity tasks. Notably, the performance decrease diminishes as model size increases, indicating that larger models might be capable of learning the intrinsic relationships and distinctions between tasks during multi-task training. This capability potentially allows these models to narrow the performance gap between multi-task and single-task training, and in some cases even achieve improvements over single-task training. This suggests that LLMs with more parameter numbers have the potential to serve as versatile general-purpose retrievers across multiple retrieval tasks.

\section{Conclusions}
In this paper, we conduct a comprehensive empirical investigation into the benefits and configurations of LLMs as backbone encoders for dense retrieval tasks. Our focus is on comparing LLMs with non-LLMs and analyzing the impact of various LLM configurations, such as parameter count, pre-training sufficiency, and alignment processes.
Our study highlights the significant advantages of utilizing LLMs as backbone encoders for dense retrieval tasks. We find that increasing the parameter count and ensuring sufficient pre-training of backbone encoders enhance in-domain accuracy. Additionally, adopting larger models consistently yields performance gains in zero-shot retrieval generalization, lengthy retrieval generalization, and multi-task learning.
These insights provide a foundation for future research aimed at optimizing dense retrieval models by balancing model size and pre-training sufficiency of backbone LLMs to achieve superior performance across diverse retrieval scenarios.

\bibliography{emnlp2023-latex/emnlp2024}
\bibliographystyle{acl_natbib}

\clearpage
\appendix

\begin{table*}[t]
    \centering
    \setlength{\tabcolsep}{28pt}
    \begin{adjustbox}{width=\textwidth}
    \begin{tabular}{lccccc}
    \toprule
    \textbf{Model} & \textbf{Dimension} & \textbf{NDCG@10} & \textbf{MRR@10} & \textbf{R@10} & \textbf{R@1000} \\
    \midrule
    BERT-base      & 768   & 37.5 & 31.6 & 57.4 & 95.2 \\
    BERT-large     & 1024  & 39.9 & 33.8 & 60.3 & 96.0 \\
    T5-base        & 768   & 40.1 & 33.7 & 61.5 & 97.3 \\
    T5-xl          & 2048  & 42.3 & 35.8 & 64.0 & 98.3 \\
    T5-xxl         & 4096  & 44.2 & 37.6 & 66.2 & 98.6 \\
    Phi-1.5-1.3B   & 2048  & 40.6 & 34.1 & 62.2 & 98.0 \\
    Phi-2-2.7B     & 2560  & 43.3 & 36.6 & 65.8 & 98.6 \\
    Gemma-2B       & 2048  & 46.8 & 39.8 & 70.1 & 99.2 \\
    Gemma-7B       & 3072  & 48.7 & 41.7 & 72.1 & 99.4 \\
    Llama-2-7B     & 4096  & 47.8 & 40.8 & 70.9 & 99.4 \\
    Llama-3-8B     & 4096  & 49.0 & 42.1 & 71.9 & 99.5 \\
    Llama-2-13B    & 5120  & 48.7 & 42.0 & 71.4 & 99.5 \\
    Qwen1.5-0.5B   & 1024  & 43.3 & 36.7 & 65.5 & 98.2 \\
    Qwen1.5-4B     & 2048  & 46.8 & 40.0 & 69.7 & 99.2 \\
    Qwen1.5-14B    & 5120  & 48.3 & 41.3 & 71.5 & 99.4 \\
    Qwen1.5-32B    & 5120  & 49.5 & 42.6 & 72.7 & 99.5 \\
    Qwen1.5-0.5B-Chat   & 1024  & 43.3 & 36.8 & 65.1 & 98.1 \\
    Qwen1.5-4B-Chat     & 2048  & 47.0 & 40.1 & 70.0 & 99.2 \\
    Qwen1.5-14B-Chat    & 5120  & 48.6 & 41.5 & 71.8 & 99.4 \\
    Llama-3-8B-Chat     & 4096  & 48.7 & 41.8 & 71.6 & 99.4 \\
    \bottomrule
    \end{tabular}
    \end{adjustbox}
    \caption{Detailed result of in-domain accuracy on MS MARCO.}
    \label{tab:accuracy}
\end{table*}

\begin{table*}[t]
    \centering
    \begin{adjustbox}{width=\textwidth}
    \begin{tabular}{lccccccccccccccc}
    \toprule
    Model & ArguAna & ClimateFEVER & DBPedia & FEVER & FiQA2018 & HotpotQA & NFCorpus & NQ & Quora & SCIDOCS & SciFact & Touche2020 & TRECCOVID & Avg \\
    \midrule
    Bert-base & 42.9 & 19.9 & 30.3 & 69.4 & 24.4 & 50.2 & 25.3 & 42.3 & 84.8 & 13.1 & 50.6 & 21.8 & 57.4 & 40.9 \\
    Bert-large & 43.1 & 21.7 & 31.9 & 68.1 & 26.4 & 51.4 & 26.7 & 46.4 & 85.7 & 13.8 & 54.7 & 20.7 & 59.2 & 42.2 \\
    t5-v1\_1-xxl & 44.0 & 24.6 & 35.2 & 63.4 & 36.1 & 57.5 & 31.4 & 50.3 & 85.1 & 15.1 & 62.0 & 22.7 & 52.9 & 44.6 \\
    Phi-v1.5-1.3B & 45.4 & 26.3 & 28.0 & 64.9 & 32.1 & 54.5 & 31.7 & 42.5 & 86.6 & 16.2 & 65.9 & 23.6 & 65.0 & 44.8 \\
    Phi-v2-2.7B & 49.4 & 31.2 & 34.4 & 70.7 & 38.4 & 62.2 & 36.5 & 50.8 & 86.9 & 18.5 & 67.2 & 23.3 & 66.1 & 48.8 \\
    Gemma-2B & 47.9 & 31.5 & 40.2 & 72.9 & 39.0 & 61.9 & 36.0 & 52.5 & 84.8 & 18.1 & 72.4 & 18.7 & 55.7 & 48.5 \\
    Gemma-7B & 49.9 & 31.3 & 42.8 & 73.5 & 44.0 & 67.3 & 38.1 & 60.4 & 86.9 & 18.7 & 74.7 & 21.5 & 58.3 & 51.2 \\
    Llama-2-7B & 48.7 & 31.2 & 44.4 & 76.2 & 42.3 & 68.1 & 36.2 & 57.3 & 86.8 & 18.3 & 73.8 & 19.6 & 47.8 & 50.0 \\
    Llama-2-13B & 57.4 & 30.7 & 43.9 & 70.4 & 45.6 & 67.7 & 37.1 & 60.9 & 85.8 & 17.7 & 74.6 & 21.8 & 55.0 & 51.4 \\
    Llama-3-8B & 56.1 & 30.8 & 41.6 & 72.7 & 41.7 & 66.0 & 35.2 & 56.4 & 85.8 & 17.8 & 74.0 & 20.6 & 56.9 & 50.4 \\
    Qwen1.5-0.5B & 46.0 & 26.6 & 32.9 & 68.1 & 31.9 & 56.6 & 29.8 & 43.4 & 84.6 & 15.8 & 65.4 & 13.5 & 54.7 & 43.8 \\
    Qwen1.5-4B & 50.2 & 30.5 & 40.5 & 72.9 & 39.4 & 63.7 & 35.4 & 54.3 & 85.3 & 17.5 & 70.8 & 18.3 & 58.6 & 49.0 \\
    Qwen1.5-14B & 56.5 & 30.1 & 43.0 & 73.4 & 45.0 & 64.4 & 36.4 & 59.3 & 85.7 & 19.3 & 74.2 & 21.9 & 60.8 & 51.5 \\
    Qwen1.5-32B & 57.5 & 31.3 & 44.5 & 75.3 & 47.9 & 68.0 & 37.1 & 59.7 & 86.0 & 18.8 & 75.6 & 24.5 & 60.3 & 52.8 \\
    \bottomrule
    \end{tabular}
    \end{adjustbox}
    \caption{Detailed result of zero-shot retrieval generalization.}
    \label{tab:performance}
\end{table*}

\begin{table*}[t]
    \centering
    \setlength{\tabcolsep}{24pt}
    \begin{adjustbox}{width=\textwidth}
    \begin{tabular}{lcccccc}
    \toprule
    \textbf{Model} & \textbf{256} & \textbf{512} & \textbf{1024} & \textbf{2048} & \textbf{4096} & \textbf{8192} \\
    \hline
    BERT-large & 18.0 & 18.1 & - & - & - & - \\
    Qwen1.5-0.5B & 17.6 & 19.0 & 20.1 & 21.1 & 37.1 & 44.9 \\
    Qwen1.5-4B & 22.8 & 23.9 & 25.4 & 27.1 & 49.1 & 54.9 \\
    Qwen1.5-7B & 24.3 & 26.4 & 27.8 & 28.2 & 52.3 & 55.9 \\
    Qwen1.5-32B & 26.9 & 28.4 & 28.7 & 30.8 & 54.8 & 59.0 \\
    Llama3-8B & 28.4 & 29.2 & 29.9 & 30.4 & 53.4 & 57.9 \\
    \bottomrule
    \end{tabular}
    \end{adjustbox}
    \caption{Detailed result of lengthy retrieval on narrativeqa with varying maximum input passage length.}
    \label{tab:Lengthy retrieval}
\end{table*}

\begin{figure*}[t]
\centering
\includegraphics[width=0.97\linewidth]{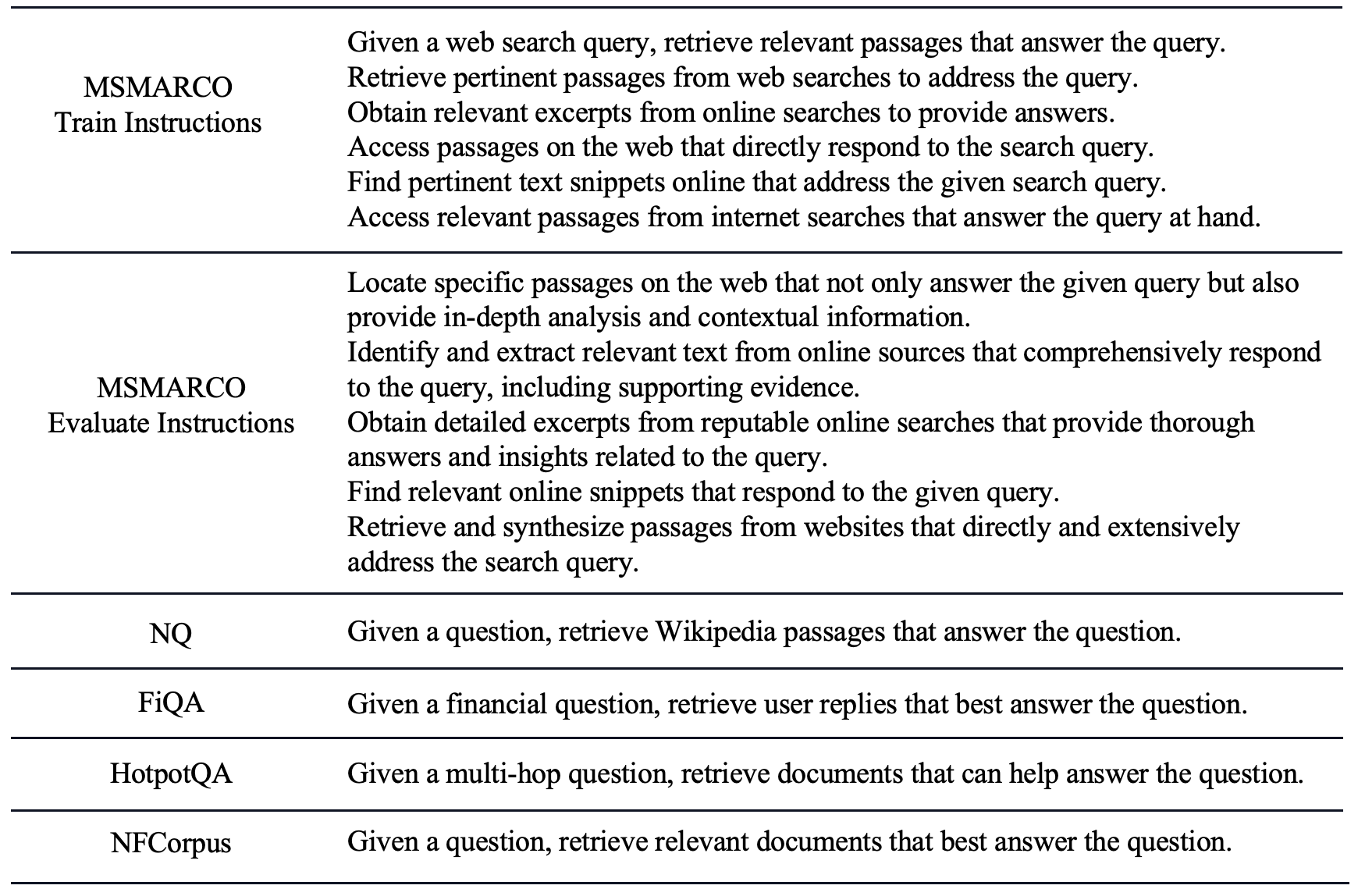}
\vspace{-9pt}
\caption{Instrctions used in instruction-based retrieval.}
\label{fig:instructions}
\end{figure*}

\end{document}